\pdfoutput=1

\documentclass[11pt]{article}

\usepackage{acl}

\usepackage{times}
\usepackage{latexsym}
\usepackage{graphicx}
\usepackage{xcolor}
\usepackage[ruled,vlined,noend,linesnumbered]{algorithm2e}
\usepackage{amsmath}
\usepackage{amsfonts}
\usepackage{booktabs}

\usepackage[T1]{fontenc}

\usepackage[utf8]{inputenc}

\usepackage{microtype}

\title{Graph-augmented Learning to Rank for Querying Large-scale Knowledge Graph}

\author{Hanning Gao\textsuperscript{1}\footnotemark[1], Lingfei Wu\textsuperscript{2}\footnotemark[1], Po Hu\textsuperscript{3}\footnotemark[2], Zhihua Wei\textsuperscript{1}\footnotemark[2], Fangli Xu\textsuperscript{4} and Bo Long\textsuperscript{5}
 \\
  \textsuperscript{1}Tongji University, \textsuperscript{2}Pinterest,
  \textsuperscript{3}Central China Normal University\\
  \textsuperscript{4}Squirrel AI Learning, \textsuperscript{5}JD.COM\\
  \texttt{gaohn@tongji.edu.cn, lwu@email.wm.edu}\\
  \texttt{phu@mail.ccnu.edu.cn, zhihua\_wei@tongji.edu.cn} \\
  \texttt{lili@yixue.us, bo.long@jd.com} \\
  }

\begin{document}
\maketitle
\renewcommand{\thefootnote}{\fnsymbol{footnote}}
\footnotetext[1]{These authors contributed equally to this work.} 
\footnotetext[2]{Corresponding authors.} 

\begin{abstract}
Knowledge graph question answering (KGQA) based on information retrieval aims to answer a question by retrieving answer from a large-scale knowledge graph. 
Most existing methods first roughly retrieve the knowledge subgraphs (KSG) that may contain candidate answer, and then search for the exact answer in the KSG.
However, the KSG may contain thousands of candidate nodes since the knowledge graph involved in querying is often of large scale, thus decreasing the performance of answer selection.
To tackle this problem, we first propose to partition the retrieved KSG to several smaller sub-KSGs via a new subgraph partition algorithm and then present a graph-augmented learning to rank model to select the top-ranked sub-KSGs from them. Our proposed model combines a novel subgraph matching networks to capture global interactions in both question and subgraphs, and an Enhanced Bilateral Multi-Perspective Matching model is proposed to capture local interactions. Finally, we apply an answer selection model on the full KSG and the top-ranked sub-KSGs respectively to validate the effectiveness of our proposed graph-augmented learning to rank method. The experimental results on multiple benchmark datasets have demonstrated the effectiveness of our approach.
\end{abstract}

\section{Introduction}

With the rise of large-scale knowledge graphs (KG) such as DBpedia \cite{auer2007dbpedia} and Freebase \cite{bollacker2008freebase}, question answering over knowledge graph has attracted massive attention recently, which aims to leverage the factual information in a KG to answer natural language question. Depending on the complexity of question, KGQA can be divided into two forms: simple and complex. Simple KGQA often requires only one hop of factual knowledge, while complex KGQA requires reasoning over a multi-hop knowledge subgraph (KSG) and selecting the correct answer among several candidate answers. In this paper, we focus on the latter, i.e., complex KGQA, which {is} more challenging.

Currently, most KGQA approaches resort to semantic parsing \cite{berant2013semantic,yih2015semantic,dong2018coarse} or retrieve-then-extract methods \cite{yao2014information,bordes2014question}.
Semantic parsing methods usually translate a natural language question to a KG query and then use it to query the KG directly. 
However, semantic parsing methods often rely on complex and specialised hand-crafted rules or schemes.
In contrast, retrieve-then-extract methods are easier to understand and more interpretable. They first retrieve the KG coarsely to obtain a KSG containing answer candidates. Then, the target answer is extracted from the retrieved KSG.
This paper follows the research idea of the retrieve-then-extract methods.

\begin{figure*}
\centering 
\includegraphics[width=0.9 \textwidth]{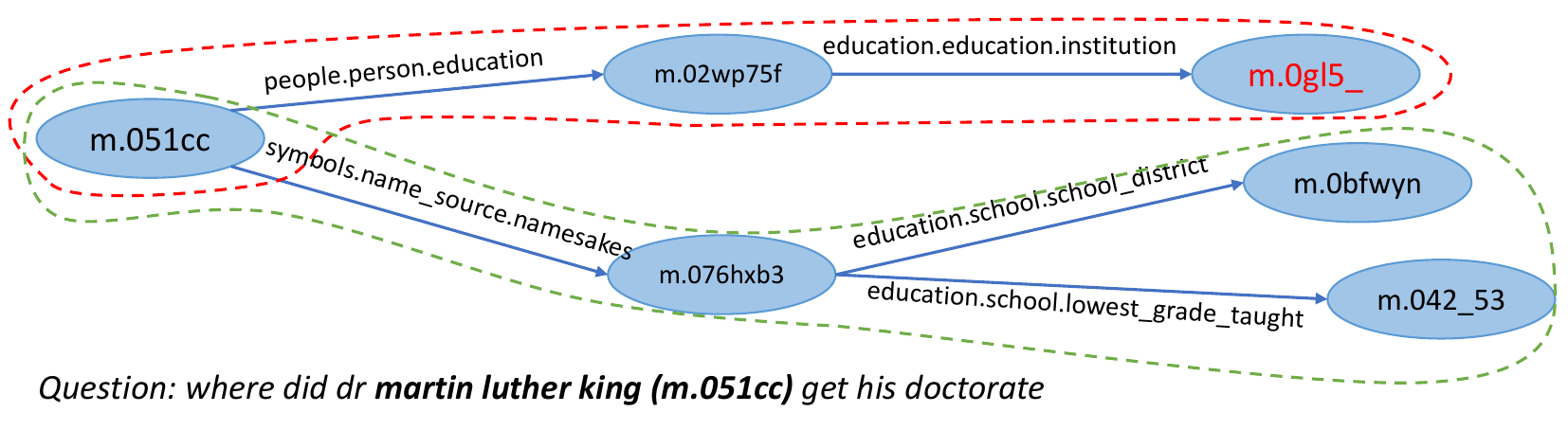} 
\caption{An Example of Knowledge Subgraph Partition Algorithm. The areas surrounded by two dashed lines belong to two different sub-KSGs.
} 
\label{fig:partition} 
\end{figure*}

Most previous works retrieve a knowledge subgraph from the original KG by choosing topic entities (e.g., KG entities mentioned in the given question) 
and their few-hop neighbors. However, since the KG is often of large volume and the initial retrieval process on it is coarse-grained and heuristic, the KSG retrieved by this method may still contain thousands of nodes and most of them are irrelevant to the given question, especially when the number of topic entities or hops significantly increases. The larger the KSG is, the more difficult it is to find the correct answer in it.
To reduce the size of the KSG, the similarity between the question and the relations around the topic entities is computed \cite{sun2018open} and then the personalized PageRank algorithm is used to select the most relevant relations. This method only considers the semantic similarity between the question and the relations while ignoring the structural information around each entity node. 
Knowledge embeddings on the whole retrieved KSG are directly computed \cite{saxena2020improving}, which is computationally intensive.

To address the above-mentioned problems, we propose a new KSG partition algorithm and a refined learning to rank model,
which focus on how to substantially reduce the size of the retrieved knowledge subgraph and ensure a high answer recall rate. 
The KSG partition algorithm is based on single source shortest path, which can partition a large-scale question-specific KSG to several moderately sized sub-KSGs. Then, the learning to rank model selects the most relevant sub-KSGs to the given question.
In this way, traditional text matching models can be used to compute the similarity score between a given question and a sub-KSG. 

However, these sequential based models often ignore the important structure information within the question and the sub-KSG. 
Therefore, we propose a novel graph-augmented learning to rank model (G-G-E) to select top-ranked sub-KSGs, which combines a novel subgraph matching networks based on Graph Neural Networks to capture global interactions between question and subgraphs, and an enhanced Bilateral Multi-Perspective Matching (BiMPM) model \cite{wang2017bilateral} to capture local interactions within parts of question and subgraphs. 
A series of graph neural networks are suitable for the subgraph matching networks  \cite{GNNBook2022}, and Gated Graph Sequence Neural Networks (GGNNs) \cite{li2015gated} is selected after comprehensive comparison.
Finally, we apply one of the state-of-the-art (SOTA) KGQA answer selection model to the original complete KSG and the merged top-ranked sub-KSGs separately, and further demonstrate that reducing the size of the answer candidate subgraphs clearly helps to select correct answer effectively and efficiently. To evaluate our approach, we conduct extensive experiments on two benchmark datasets. The experimental results on the datasets have shown that our proposed model can significantly improve subgraph ranking performance compared to existing SOTA methods.

In summary, the contributions of this paper can be summarized as follows:
\begin{itemize}
    \item We propose a new knowledge subgraph partition algorithm based on single source shortest path.
    \item We propose a novel graph-augmented learning to rank model, which combines a novel subgraph matching networks based on GGNNs and an enhanced BiMPM model.
    \item Our proposed graph-augmented learning to rank model outperforms a set of SOTA ranking models. 
    \item Further answer selection experiments on the original complete KSG and the merged top-ranked sub-KSGs demonstrate reducing the size of the answer candidate subgraphs can help improve the performance of answer selection.
\end{itemize}

\section{Knowledge Subgraph Partition}
\label{Parition-Algorithm} 
For better use of the ranking model, we need to partition the knowledge subgraph into several sub-KSGs. 
As shown in {Figure} \ref{fig:partition}, \texttt{m.051cc} is the topic entity of the given question and nodes on the same path from topic entity node \texttt{m.051cc} should be partitioned in the same sub-KSG. In particular, entity nodes in this example graph are denoted by Freebase IDs.
The first sub-KSG (the red dashed line area) is about the education information of \texttt{m.051cc}, which contains the true answer entity node \texttt{m.0gl5\_}. The second sub-KSG (the green dashed line area) is about the namesake entity \texttt{m.076hxb3}. It is also a confusing subgraph because it contains tokens like \textit{education}, which are consistent with the context of the question. Therefore, the learning to rank model is expected to distinguish not only irrelevant sub-KSGs, but also confusing ones.

\begin{algorithm}
\SetAlgoLined
\textbf{Input:} Question $q$ with its KSG $S$, topic entity $n_t$, answer entities $E_a^q$\\
Find the shortest paths $P$ to all nodes with $n_t$ as the source node; \\
Define $Set_S = \{\}$ to save all partitioned sub-KSGs; \\
Define $Set_l = \{\}$ to save the match labels of the partitioned sub-KSGs; \\
 \For{each path $p_i$ ($n_i$ as target node) in $P$}{
 \If{$n_i$ has child nodes \text{and} the child nodes of $n_i$ are all leaf nodes}{
 Partition the path from $n_t$ to $n_i$ as a sub-KSG $S_{n_i}$; \\
 Add the child nodes of $n_i$ to $S_{n_i}$ and set its match label $l_{n_i}$ as 0;\\
 \For{$n_a$ in $E_a^q$}{\If{exists path from $n_t$ to $n_a$}{Set the match label $l_{n_i}$ as 1; break;}}
  Add $l_{n_i}$ to $Set_l$ and $S_{n_i}$ to $Set_S$ ;\\
 }
 }
 \caption{KSG Partition}
 \label{alg:algorithm1}
\end{algorithm}

\begin{figure}
\centering 
\includegraphics[width=0.5 \textwidth]{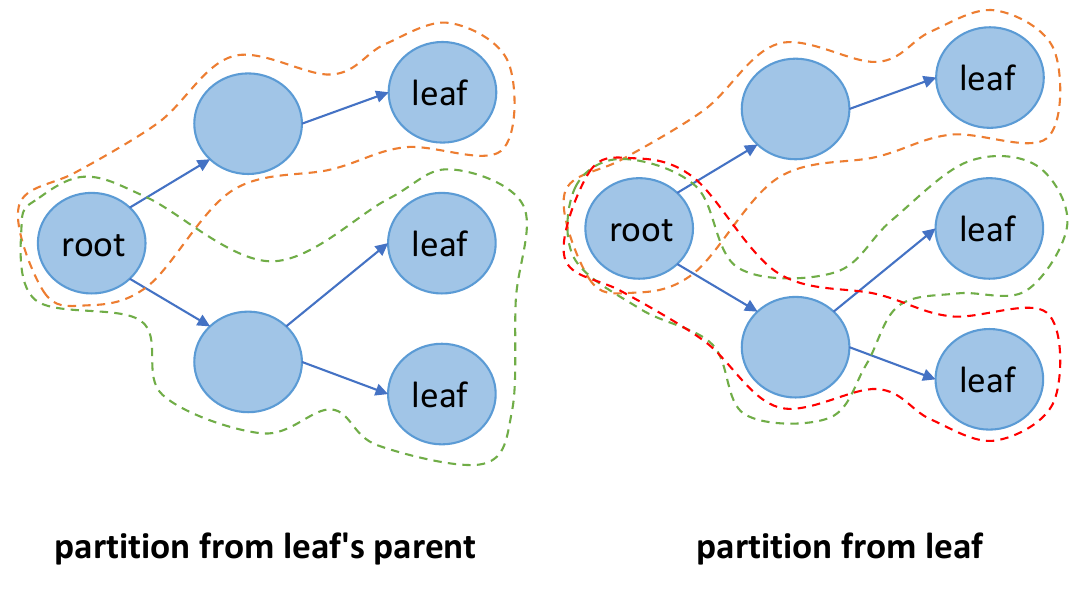} 
\caption{An example of two KSG partition methods: from the parent node whose child nodes are all leaf nodes and leaf node respectively.
} 
\label{fig:part_method} 
\end{figure}

To partition related nodes in the same sub-KSG, we propose a knowledge subgraph partition algorithm detailed in Algorithm \ref{alg:algorithm1}. Given a question $q$ and its answer entities $E_a^q$, we first use the retrieval method proposed by \cite{sun2018open} to obtain a question-specific KSG $S$, which may contain thousands of answer candidate entities and relationships. $E_a^q$ is a set containing the ground truth answer entities for question $q$.
Then, our proposed algorithm partitions the retrieved KSG into several sub-KSGs serving as inputs to the graph-augmented learning to rank model to select the most relevant sub-KSGs. 
Our algorithm follows the intuition that the answer to the given question is usually found on a multi-hop path from the topic entity node. In order to keep the size of the sub-KSG moderate, we partition it from the node whose child nodes are all leaf nodes, which is shown in the left of Figure \ref{fig:part_method}. The reason for partitioning from such nodes is two-fold. Firstly, if partitioned from a leaf node (see the right of Figure \ref{fig:part_method}), the sub-KSG will degrade to a sequence and the number of sub-KSGs will be too large. Second, if partitioned from a parent node near the root node, the sub-KSG may still contain too much redundant information for a given question.

\section{Graph-augmented Learning to Rank}

\begin{figure*}
\centering 
\includegraphics[width=0.7\textwidth]{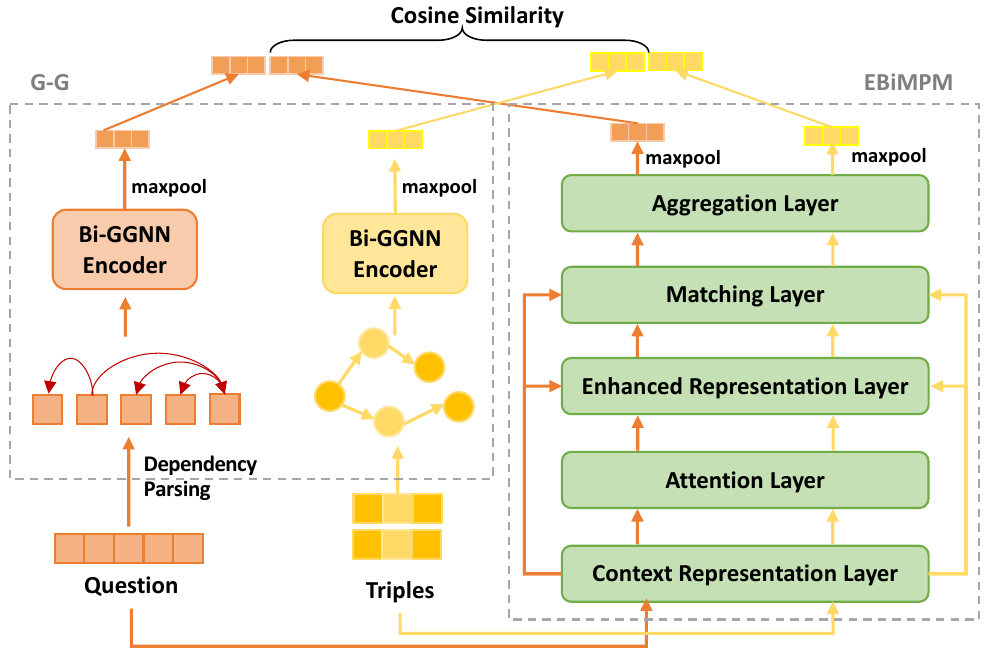}
\caption{The Proposed G-G-E Model Architecture. The model contains two components: (1) A Subgraph Matching Networks component on the left (i.e., G-G in the figure); (2) An Enhanced BiMPM component on the right (i.e., EBiMPM in the figure).
} 
\label{fig:model} 
\end{figure*}

Given a question $q$ and a set of sub-KSGs $S_q = \{S_{q,1}, ..., S_{q,n}\}$, we compute the ranking score $y$ representing the relevance of $q$ and $S_{q,i}$ for subgraph ranking. The overall model architecture is shown in Figure \ref{fig:model}, which consists of a graph construction module for the input question and the input triples, a BiGGNN encoder and an Enhanced BiMPM encoder.

\subsection{Graph Constructions}
\paragraph{Question Graph.} Question graph $G_q$ is a directed graph constructed by the dependency parser from Stanford CoreNLP \cite{manning2014stanford}.
The dependency parsing graph represents the grammatical structure of the input question. Nodes in the dependency parsing graph are the tokens in the question and an edge indicates a modified relationship between two token nodes. In particular, we only use the connection information for the edges, not the labels for the edges.

\paragraph{Sub-Knowledge Subgraph.} A sub-KSG consists of a set of triples $S_{q,i} = \{(s, r, o)| \\ s, o \in \mathcal{E}, r \in \mathcal{R}\}$, where $\mathcal{E}$ and $\mathcal{R}$ denote the entity and relation set. Relation $r$ is regarded as an additional node. We assume there is a directed edge from subject node $s$ to $r$, and another directed edge from $r$ to subject node $o$. In the following sections, we will introduce how to calculate a relevant score between a question $q$ and a subgraph $S_{q,i}$ ($S$ for short).

\subsection{Subgraph Matching Networks}
To better exploit the global contextual information and the structural information, we expand GGNNs from uni-directional to bi-directional. Given a question graph $G_q$ or a sub-KSG $S$, each node $v$ is initialized with its word embedding (e.g., average word embeddings for multi-token nodes).
To calculate the representation of each node $\mathbf{h}_v^{(l)}$ at layer $l$, the encoder first aggregates the information of neighbouring nodes to compute aggregation vectors using the following update rule:
\begin{align}
    \mathbf{m}_{v\vdash}^{(l)}=\sum_{{u} \in N_{\vdash} (v)} \mathbf{W}_{\vdash}^{(l-1)} \mathbf{h}_{u\vdash}^{(l-1)} \\
    \mathbf{m}_{v\dashv}^{(l)}=\sum_{{u} \in N_{\dashv} (v)} \mathbf{W}_{\dashv}^{(l-1)} \mathbf{h}_{u\dashv}^{(l-1)}
\end{align}
where $N_{\vdash}(v)$ and $N_{\dashv}(v)$ denote the neighbours of $v$ with outgoing and ingoing edges. $\mathbf{W}^{(l-1)}_{\vdash}$ and $\mathbf{W}^{(l-1)}_{\dashv}$ are trainable weight matrices. Then, a Gated Recurrent Unit (GRU) \cite{cho2014learning} is used to update the node representation at layer $l$ based on the aggregation vectors and the node representation at previous layer:
\begin{align}
\mathbf{h}_{v\vdash}^{(l)}=\text{GRU}(\mathbf{m}_{v\vdash}^{(l)}, \mathbf{h}_{v\vdash}^{(l-1)})  \\
\mathbf{h}_{v\dashv }^{(l)}=\text{GRU}(\mathbf{m}_{v\dashv}^{(l)}, \mathbf{h}_{v\dashv}^{(l-1)})
\end{align}
After obtaining all node representations of an input graph, max pooling is applied to compute the graph embedding:
\begin{equation}
\label{equ:ge}
    \mathbf{r} = \text{max}(\{[ \mathbf{h}_{v\vdash}^{(L)}; \mathbf{h}_{v\dashv}^{(L)}], \forall v \in \mathcal{N}\})
\end{equation}
where $\mathcal{N}$ is the node set and $L$ is the maximum number of layers. $\mathbf{r}_q$ is the question graph embedding and $\mathbf{r}_S$ is the sub-KSG graph embedding.
The concatenation representation of node $v$ is $[ \mathbf{h}_{v\vdash}^{(L)}; \mathbf{h}_{v\dashv}^{(L)}] \in \mathbb{R}^{2D}$ and the set of node representations is in $\left | \mathcal{N} \right | \times 2D$ dimension. The max pooling operation is applied on the first dimension and the graph embedding is $\mathbf{r} \in \mathbb{R}^{2D}$.

\begin{table*}
\centering
\begin{tabular}{c|c|c|c|c|c|c}
\toprule[1pt]
Dataset & \# Train & \# Dev & \# Test & \# Entities in KSG & \# Sub-KSGs & Coverage Rate \\ \hline
WebQSP  &  2848 & 250 & 1639 & 1429.8 & 1279.9 &  94.9\% \\
CWQ     & 18391 & 2299 & 2299 & 95.9 & 50 & 95.7\% \\ 
\toprule[1pt]
\end{tabular}
\caption{Statistics information of the WebQSP dataset and the CWQ dataset.}
\label{Statistics-table}
\end{table*}

\subsection{Enhanced BiMPM}
Bilateral Multi-Perspective Matching (BiMPM) is a strong text matching model due to its capacity of capturing the local interactions.
To better learn local interactions for sentence between the question and the sub-KSG, we propose to add an attention layer and an enhanced representation layer on the basis of the original BiMPM model.
Specifically, our proposed EBiMPM first uses a shared BiLSTM-based context representation layer to encode two input sequences to get two embeddings $\mathbf{q} \in \mathbb{R}^{l_1 \times d}$ and $\mathbf{S} \in \mathbb{R}^{l_2 \times d}$, where $l_1$ and $l_2$ are the lengths of the input texts.
Second, the newly-added attention layer applies a bi-directional attention mechanism between $\mathbf{q}$ and $\mathbf{S}$. The attentive embedding of the i-th question token $\mathbf{q}_i$ over $\mathbf{S}$ is computed as:
\begin{equation}
    \widetilde{\mathbf{q}_i} = \sum_{j=1}^{l_2} \frac{\text{exp}(\mathbf{q}_i^T\mathbf{S}_j)} {\sum_{k=1}^{l_2} \text{exp}(\mathbf{q}_i^T\mathbf{S}_k)} \mathbf{S}_j
\end{equation}
Similarly, we can compute the attentive embedding $\mathbf{\widetilde{S_i}}$ of the i-th sub-KSG token $\mathbf{S}_i$ over $\mathbf{q}$:
\begin{equation}
    \widetilde{\mathbf{S}_i} = \sum_{j=1}^{l_1} \frac{\text{exp}(\mathbf{S}_i^T\mathbf{q}_j)} {\sum_{k=1}^{l_1} \text{exp}(\mathbf{S}_i^T\mathbf{q}_k)} \mathbf{q}_j
\end{equation}

The attention layer outputs the attentive embeddings $\mathbf{\widetilde{q}}$ and $\mathbf{\widetilde{S}}$. Third, the enhanced representation layer fuses $\mathbf{q}$ and $\mathbf{\widetilde{q}}$ using:
\begin{equation}
\widehat{\mathbf{q}} = f([\mathbf{q};\widetilde{\mathbf{q}};\mathbf{q}-\widetilde{\mathbf{q}};\mathbf{q}\odot  \widetilde{\mathbf{q}}])
\end{equation}
where $f(\cdot)$ is a one-layer perceptron and $\odot$ is the point-wise multiplication operation. We can also compute the enhanced subgraph representation $\widehat{\mathbf{S}}$.

Then, $\mathbf{q}$ and $\mathbf{S}$ are fed into the BiMPM matching layer \cite{wang2017bilateral} to get two sequences of matching vectors $\overline{\mathbf{q}} \in \mathbb{R}^{l_1 \times 8l}$ and $\overline{\mathbf{S}} \in \mathbb{R}^{l_2 \times 8l}$, where $l$ is the number of perspectives.
For the matching layer, we follow the original implementation of BiMPM, which defines four kinds of matching strategies to compare each time-step of one sequence against all time-steps of the other sequence from both forward and backward directions.

Finally, [$\overline{\mathbf{q}}$;$\widehat{\mathbf{q}}$] and [$\overline{\mathbf{S}}$;$\widehat{\mathbf{S}}$] are regarded as inputs to a shared BiLSTM-based aggregation layer to get the final representation:
\begin{align}
    \mathbf{r}_q' = \text{max}(g([ \overline{\mathbf{q}};\widehat{\mathbf{q}}]))
    \ \ \text{and} \ \
    \mathbf{r}_S' = \text{max}(g([\overline{\mathbf{S}};\widehat{\mathbf{S}}]))
\end{align}
where $\text{max}(\cdot)$ is max pooling and $g(\cdot)$ is a BiLSTM aggregation layer.

\subsection{Ranking Score Function}
The representations of the question and the sub-KSG learned by the subgraph matching networks and EBiMPM are concatenated separately and input to a cosine similarity ranking score function:
\begin{equation}
    \hat{y} = cos([\mathbf{r}_q;\mathbf{r}_q'], [\mathbf{r}_S;\mathbf{r}_S'])
\end{equation}
At last, we take Mean Square Error (MSE) as the loss function:
\begin{equation}
    L = \frac{1}{N_m}\sum_{m=1}^{N_m}(y_m-\widehat{y_m})^2
\end{equation}
where $N_m$ is the number of samples and $y_m$ is the label. 

\subsection{Answer Selection Model}
After using the ranking model to obtain the top sub-KSGs, we merge them into a smaller graph compared to the original large KG graph and feed it into an answer selection model. In this paper, we use one of the state-of-the-art KGQA model GraftNet \cite{sun2018open} as our answer selection model, which is a heterogeneous graph neural network model. To improve the overall performance, GraftNet also incorporates external Wikipedia knowledge and computes a PageRank \cite{haveliwala2003topic} score for each entity node. However, we only use the basic model of GraftNet as our answer selection model to better validate the effectiveness of our proposed graph-augmented learning to rank model. GraftNet performs a binary classification to select the answer:
\begin{equation}
    Pr(v|q, S) = \sigma (\mathbf{W} \mathbf{h}_v^{(L)}+\mathbf{b})
\end{equation}
where $\mathbf{h}_v^{(L)}$ is the final nodes representation learned by GraftNet and $\sigma$ is the sigmoid function. This model is trained with binary cross-entropy loss, using the full KSG and the merged top-ranked sub-KSGs as input respectively.

\section{Experiments}

\subsection{Datasets}
\label{sec: Dataset}

\begin{table*}[]
\centering
\begin{tabular}{l|cccccc|cccc}
\toprule[1pt]
Dataset & \multicolumn{6}{c|}{WebQSP} & \multicolumn{4}{c}{CWQ} \\ \hline
Model &  MRR &  R@1 &  R@10 &  R@100 &  R@200 &  R@300   &  MRR  & R@1 & R@10 & R@20 \\ \hline
BiMPM & 0.612 & 0.531 & 0.766 & 0.882 & 0.903 &	0.912 & 0.680 & 0.570 & 0.906 & 0.965 \\ 
EBiMPM & 0.661 &  0.595 &  0.780 & 0.880 & 0.899 &	0.909 & 0.707 & 0.609 & 0.906 & 0.964 \\ 
BERT & 0.682 & 0.619 & 0.789 & 0.885 & 0.905 & 0.914 & 0.736 & 0.664 & 0.884 & 0.951 \\ 
G-G & 0.687 & 0.632 & 0.790 & 0.880 & 0.905 & 0.918 & 0.712 & 0.637 & 0.871 & 0.940 \\
\textbf{G-G-E} &  \textbf{0.698} & \textbf{0.643} &  \textbf{0.797} & \textbf{0.891} & \textbf{0.913} & \textbf{0.924} & \textbf{0.754} & \textbf{0.675} & \textbf{0.923} & \textbf{0.967}  \\ 
\toprule[1pt]
\end{tabular}
\caption{Ranking Experimental Results. Bold fonts indicate the best results.}
\label{results-table}
\end{table*}

We {conduct} experiments on two multi-hop question answering datasets, i.e., WebQuestionsSP (WebQSP) \cite{yih2015semantic} and ComplexWebQuestions (CWQ) \cite{talmor2018web}.
Table \ref{Statistics-table} shows the statistical information of the datasets. 
For WebQSP, we use the partition algorithm to construct the sub-KSGs based on the processed data \cite{He-WSDM-2021}, which follows the retrieval method proposed in \cite{sun2018open}. Because the dataset is small, the train and dev matching datasets used for training phase are constructed by selecting a sub-KSG containing true answers and random sampling 20 sub-KSGs for each example. For the test dataset, each example contains a natural language question and all partitioned sub-KSGs. The model computes a ranking score for each (question, sub-KSG) pair. The average number of entities in each KSG is 1429.9 and each KSG produces an average of 1279.9 sub-KSGs after the partition process. The coverage rate, namely the percentage of examples that can find answers in their corresponding KSGs, is 94.9\%.

For CWQ, we use the preprocessed datasets released by \cite{kumar2019difficulty}. 
Each sample contains a question, a subgraph from which the question is derived and a set of answer entities. 
The CWQ dataset contains 22989 matched (question, subgraph) pairs. The division ratio of train set, dev set and test set is 8:1:1.
For the train set and the dev set, we produce the same number of negative examples as the positive ones. For each question, we select a confusion-prone subgraph from the training subgraph set that is similar to the matched subgraph but contains no answer nodes as a negative sample. TF-IDF is used to compute the similarity of the text of two subgraphs. 
For the test dataset used for ranking evaluation, it consists of a matched subgraph and 49 unmatched subgraphs which are similar to the matched one. Therefore, the average number of sub-KSG (subgraph) for the CWQ dataset is 50. We merge these 50 sub-KSGs (subgraphs) to form a pseudo KSG for each example. 
The average number of entities in a pseudo KSG is 95.9 and the coverage rate of the test dataset is 95.7\%.

\subsection{Models and Metrics}
In the next experiments, our proposed BiGGNN-BiGGNN-EBiMPM (G-G-E) model is compared with the following baselines: 
\begin{itemize}
    \item {BiMPM \cite{wang2017bilateral}:} an LSTM-based model for text matching;
    \item {EBiMPM:} BiMPM with an attention layer and an enhanced representation layer;
    \item {BERT \cite{devlin2018bert}:} a shared BERT model to encode the question sequence and the subgraph triples sequence;
    \item {BiGGNN-BiGGNN (G-G):} both question graph and sub-KSG are encoded by a BiGGNN respectively;
\end{itemize}

To evaluate the graph-augmented learning to rank model, we use Recall@K (R@K) and Mean Reciprocal Rank (MRR) as the evaluation metrics. Recall@K is the proportion of examples that can find sub-KSGs containing answers in the top-K sub-KSGs. Mean reciprocal rank is the average of the reciprocal ranks of the sub-KSGs containing answers. 
Furthermore, we use Hits, precision, recall and F1 to evaluate whether reducing the size of the KSG is beneficial to the subsequent answer selection model. Hits is the proportion of examples where GraftNet can select answer nodes in the subgraph merging the top-K sub-KSGs.

\subsection{Experimental Settings}
Our proposed model are implemented by MatchZoo-py \cite{Guo:2019:MLP:3331184.3331403} and Graph4NLP \cite{wu2021graph}. We use Adam \cite{kingma2014adam} optimization with an initial learning rate 0.0005. The batch size is 64 for CWQ and is 50 for WebQSP. Word embeddings are initialized with 300-dimensional pretrained GloVe \cite{pennington2014glove} embeddings
. BiGGNN encoder is stacked to 2-layer.  Early stopping is introduced during the training phase and the validation set is evaluated every epoch. 
All models use cosine similarity as ranking score function. All experiments are run on Tesla V100.

\subsection{Results Analysis}

\begin{table*}[]
\centering
\begin{tabular}{c|cccc|c|cccc}
\toprule[1pt]
Dataset & \multicolumn{4}{c|}{WebQSP}   & \multicolumn{5}{c}{CWQ}   \\ \hline
Data    & Hits  & Precision                     & Recall & F1    & Data   & Hits  & Precision   & Recall & F1    \\ \hline
top 100 & 0.604 & 0.604 & 0.582 & 0.513 & top 10 & \textbf{0.424} & 0.530&  \textbf{0.411} & \textbf{0.327} \\
top 200 & 0.598 & \textbf{0.656} & 0.586 & 0.536 & top 20 & 0.400   & 0.515 & 0.377 & 0.292 \\
top 300 & \textbf{0.605} & 0.620  & \textbf{0.639} & \textbf{0.550} & full & 0.396 & \textbf{0.567}& 0.339 & 0.274\\
full    & 0.579 & 0.574 & 0.625 & 0.522 &        &       &       &       & \\ 
\toprule[1pt]
\end{tabular}
\caption{Answer selection results on WebQSP and CWQ.}
\label{answer-selection}
\end{table*}

\begin{table*}
\centering
\begin{tabular}{l}
\toprule[1pt]
\textbf{Question:} what artistic movement did \texttt{m.0gct\_} belong to ? \\ \hline
\textbf{M:}(\texttt{m.0gct\_} , influence\_influence\_node\_influenced\_by,  \texttt{m.0160zv}) \\ (\texttt{m.0160zv},  visual\_art\_visual\_artist\_associated\_periods\_or\_movements , \texttt{m.0160zb}) \\
\textbf{R:}(\texttt{m.0gct\_}, visual\_art\_visual\_artist\_associated\_periods\_or\_movements, \texttt{m.049xrv}) \\ \hline
\textbf{Question:} who did \texttt{m.01ps2h8} play in lord of the rings ? \\ \hline
\textbf{M:}(\texttt{m.01ps2h8}, film\_actor\_film, \texttt{m.0k5s9k}), (\texttt{m.0k5s9k}, film\_performance\_film, \texttt{m.017gl1}) \\
\textbf{R:}(\texttt{m.01ps2h8}, film\_actor\_film \texttt{m.0k5sfk}), (\texttt{m.0k5sfk}, film\_performance\_character, \\ \texttt{m.0gwlg}) \\ \hline
\toprule[1pt]
\end{tabular}
\caption{An example of mispredicted subgraph by our model on the WebQSP dataset. M and R denote Mispredicted and Real respectively. }
\label{case-table} 
\end{table*}

Table \ref{results-table} shows the ranking performance on two datasets. In particular, the upper limit of Recall@K is 100\% rather than the coverage rate because we eliminate examples for which we can not find an answer.
It can be seen that our proposed full model G-G-E consistently outperforms other baselines on all datasets, including the BERT model. To guarantee a high answer recall for the merged subgraph, we are more concerned about Recall@K than Recall@1, especially when K is large. Our proposed G-G-E model is 0.6 to 1 percentage point higher than the best baseline models for metrics Recall@100, Recall@200 and Recall@300 in dataset WebQSP. In the dataset CWQ, the Recall@10 of the G-G-E model {is also improved by} 1.7\% compared to the best baseline model.
Moreover, on the WebQSP dataset, G-G is significantly better than BiMPM, increasing by 0.07 on MRR and 0.1 on Recall@1 respectively, which indicates the graph structure information plays a more important role on this dataset. 


To further validate that reducing the size of KSG helps improve the performance of answer selection, we merge the top 100, 200 and 300 sub-KSGs of the WebQSP dataset and the top 10, 20 sub-KSGs of the CWQ dataset. The experimental results are shown in Table \ref{answer-selection}. For WebQSP, the answer selection model performs best on the top-300 merged subgraph, increasing by 0.026 on Hits and 0.027 on F1. The top-300 merged subgraph is almost a third of the size of the original full KSG, which contains an average of 1280 sub-KSGs. The improvements also verify the effectiveness of our proposed partition algorithm. For CWQ, the answer selection model performs best on the top-10 merged subgraph, increasing by 2.8\% on Hits and 5.4\% on F1. The top-10 merged subgraph is a fifth of the size of the full KSG. 
From the above two results we can see that the answer selection model performs better on the subgraph merging the top-K relevant sub-KSGs than on the full KSG. This is because the answer selection model is easier to find the correct answer entity node in a graph that contains fewer noisy nodes.
In general, by using our proposed partition algorithm and graph-augmented learning to rank model, we can further reduce the size of the KSG, while ensuring the answer recall rate.


\subsection{Ablation Study and Case Study}
We conduct an ablation study to investigate the contribution of each component to the proposed model. As shown in Table \ref{results-table}, we evaluate models with only graph neural network encoder (G-G) and with only sequence encoder (EBiMPM), respectively. The performance gain of G-G-E model compared to G-G and EBiMPM can empirically demonstrate the effectiveness of combining the two encoders for capturing both global and local interactions between the question and the knowledge subgraph.

Furthermore, we manually {check the sub-KSGs that are} incorrectly considered as containing answers to study the limitations of our proposed model. The topic entity in the question and the entities in the subgraph are replaced by their Freebase ID. 
As shown in {Table \ref{case-table}}, the first mispredicted subgraph contains a redundant hop ``influence\_influence\_node\_influenced\_by''. This may because our model ignores the number of hops of the question. 
The second example fails to map \textit{play} in the question to the relation \textit{film\_performance\_character}. 
It confuses the model because the mispredicted subgraph is very similar to the real one.

\section{Related Work}
\subsection{Knowledge Graph Question Answering}
With the rapid development of large-scale knowledge graphs (KG) such as DBpedia \cite{auer2007dbpedia} and Freebase \cite{bollacker2008freebase}, question answering over knowledge graph has attracted widespread attention from a growing number of researchers. However, due to the large volume of the knowledge graph, using the knowledge in it to answer questions is a challenging task. 
Knowledge Graph Question Answering has two mainstream research methods, namely semantic parsing based methods and retrieve-then-extract methods. 

\paragraph{Semantic parsing based methods} 
convert natural language questions to knowledge base readable queries, which can be summarised in the following steps \cite{lan2021survey}: (1) Using a \textit{Question Understanding} module to analyze questions semantically and syntactically. Common question analysis techniques include dependency parsing \cite{abujabal2017automated}, AMR parsing \cite{kapanipathi2021leveraging} and skeleton parsing \cite{sun2020sparqa}. (2) Using a \textit{Logical Parsing} module to convert the question embedding into an uninstantiated logic form. This module creates a syntactic representation of the question such as template based queries \cite{bast2015more} and query graphs \cite{hu2018state}. (3) Using a \textit{KB Grounding} module to align the logic form to KB \cite{bhutani2019learning, chen2019uhop}. The logical query obtained from the above steps can be searched directly in KB to find the final answer. 

\paragraph{Retrieve-then-extract methods}
are also known as information retrieval based methods. 
A subgraph retrieval method and a subgraph embedding model which can score every candidate answer were first proposed in \cite{bordes2014question}. In the following work, a memory table was adopted to store KB facts encoded into key-value pairs \cite{miller2016key}. A graph neural network model was proposed in \cite{sun2018open} to perform multi-hop reasoning on heterogeneous graphs. PullNet \cite{sun2019pullnet} improved the graph retrieval module by iteratively expanding the question-specific subgraph. BAMnet \cite{chen2019bidirectional} modeled the bidirectional flow of interactions between the questions and the KB using an attentive memory network. EmbedKGQA \cite{saxena2020improving} directly matched pretrained entity KG embeddings with question embedding, which is computationally intensive.

\subsection{Learning to Rank}

Traditional learning to rank models rely on hand-crafted features, which are often time-consuming to design.
Recently, many ranking models based on neural networks have emerged. Deep Structured Semantic Model (DSSM) \cite{huang2013learning} is the first neural network ranking model using fully connected neural networks. 
A match-LSTM model combining Pointer Net \cite{vinyals2015pointer} is proposed in \cite{wang2016machine}. 
ANMM \cite{yang2016anmm} is an attention based neural matching model combining different matching signals for ranking short answer text. BiMPM \cite{wang2017bilateral} uses the \textit{matching-aggregation} framework to match the sentences from multiple perspectives. With the development of pretrained language models such as BERT \cite{devlin2018bert}, the performance of neural ranking models is taken to a next level. These neural ranking models have limitations when applied to information retrieval based KGQA because the inputs are considered as raw text sequences and the structural information in the KG is ignored.

\section{Conclusions}
In the information retrieval based {Knowledge Graph Question Answering (KGQA)}, this paper focuses on a subgraph ranking task with the aim of reducing the size of the coarsely retrieved knowledge subgraph and capturing both local and global interactions between question and sub-KSGs. We propose a {knowledge subgraphs (KSG)} partition algorithm and a graph-augmented learning to rank model to match-then-rank them. We further validate that reducing the size of knowledge subgraph is beneficial to the subsequent answer selection in an information retrieval based KGQA process. 
In the future, we will further explore a more effective answer selection model over the small-scale knowledge subgraph selected by our learning to rank model.

\section*{Acknowledgements}
The work is partially supported by the National Nature Science Foundation of China (No. 61976160), the Research Project of State Language Commission (No. YB135-149) and the Self-determined Research Funds of CCNU from the Colleges' Basic Research and Operation of MOE (No. CCNU20ZT012, CCNU20TD001).

\section*{Ethical Considerations}
In the ethical context of our work, it is important to consider real-world use cases, impacts, and potential users. The primary real-world application of our methods is in question answering systems or knowledge-enhanced retrieval applications, where our model and relevant techniques could be used to improve question-understanding and response or information accessing ability of such systems.
However, we do not yet prepare our current trained models to be employed immediately in such real-world applications, given that our models were just trained and tested on a few benchmark datasets which are widely used for KGQA task. More complicated real-world applications built on our work should be re-trained using one or more task-oriented training datasets, because our model has not tuned for any specific application scenario. 
Our methods could also be used in diverse contexts e.g. education or health-care settings, and it is essential that any such applications undertake quality-assurance and robustness testing, as our solution is not designed to meet stringent robustness requirements (e.g., for not stating false facts or meeting legal requirements).
More generally, there is the possibility of (potentially harmful) social biases that can be introduced in training data. Again, we would urge potential users to undertake the necessary testing to evaluate the extent to which such biases might be present and impacting their trained system.


\bibliography{anthology,custom}

\appendix



\end{document}